\documentclass[runningheads]{llncs}
\usepackage{graphicx}
\usepackage[english]{babel}

\usepackage[letterpaper,top=2cm,bottom=2cm,left=3cm,right=3cm,marginparwidth=1.75cm]{geometry}

\usepackage{amsmath}
\usepackage{amssymb}
\usepackage{graphicx}
\usepackage{authblk}
\usepackage[colorlinks=true, allcolors=blue]{hyperref}
\usepackage{subfig}

\title{ICU Bloodstream Infection Prediction: A Transformer-Based Approach for EHR Analysis}

\author{Ortal Hirszowicz\inst{1} \and
Dvir Aran\inst{1,2}\orcidID{0000-0001-6334-5039}}
\authorrunning{O. Hirszowicz and D. Aran}
\institute{Taub Faculty of Computer Science, Technion-Israel Institute of Technology, Haifa, Israel \\ Faculty of Biology, Technion-Israel Institute of Technology, Haifa, Israel\\ \email{dviraran@technion.ac.il}}

\begin{document}
\maketitle

\begin{abstract}

We introduce RatchetEHR, a novel transformer-based framework designed for the predictive analysis of electronic health records (EHR) data in intensive care unit (ICU) settings, with a specific focus on bloodstream infection (BSI) prediction. Leveraging the MIMIC-IV dataset, RatchetEHR demonstrates superior predictive performance compared to other methods, including RNN, LSTM, and XGBoost, particularly due to its advanced handling of sequential and temporal EHR data. A key innovation in RatchetEHR is the integration of the Graph Convolutional Transformer (GCT) component, which significantly enhances the ability to identify hidden structural relationships within EHR data, resulting in more accurate clinical predictions. Through SHAP value analysis, we provide insights into influential features for BSI prediction. RatchetEHR integrates multiple advancements in deep learning which together provide accurate predictions even with a relatively small sample size and highly imbalanced dataset. This study contributes to medical informatics by showcasing the application of advanced AI techniques in healthcare and sets a foundation for further research to optimize these capabilities in EHR data analysis.

\end{abstract}


\section{Introduction}

The advent of Electronic Health Records (EHRs) has revolutionized the landscape of healthcare data management, offering unprecedented opportunities for enhancing patient care and clinical decision-making. Particularly in Intensive Care Units (ICUs), where patients are at high risk and require close monitoring, the effective analysis of EHRs can be a lifesaver. One of the most critical applications of EHR analysis in ICU settings is the early detection of bloodstream infections (BSIs), a condition associated with high morbidity and mortality rates \cite{10.1001/archinte.1995.00430060050006}. Traditional approaches to predicting BSIs have relied on a variety of statistical and machine-learning approaches, yet these have often fallen short due to the complex, temporal, and high-dimensional nature of EHR data.

Existing models for BSI prediction \cite{roimi,zoabi2021,zhang,dongchoi} have faced significant challenges in capturing the nuanced temporal dynamics and intricate feature inter-dependencies present in EHR data. These limitations stem primarily from the challenges in processing sequential data effectively, leading to a loss of critical information. Furthermore, the high dimensionality and sparsity of EHR data, coupled with issues like class imbalance and the need for extensive data preprocessing, have further complicated the predictive analysis.

In response to these challenges, we introduce RatchetEHR, leveraging the power of transformer-based architecture \cite{vaswani2017attention}, to analyze ICU EHRs for BSI prediction. This approach allows to effectively capture the sequential nature and hidden structures within the data. We show that this approach not only enhances prediction accuracy but also offers a deeper understanding of the underlying patterns and relationships in EHR data.

\section{Related Work}

The transformer architecture \cite{vaswani2017attention} emerges as a particularly fitting model for modeling EHR data due to several intrinsic properties of EHR and the strengths of transformers. EHR data is inherently complex, comprising long sequences of patient visits, each containing various medical elements like diagnoses, treatments, and observations. This complexity and sequential nature of EHR data align well with the capabilities of transformers. Transformers excel in handling sequential data, a property leveraged extensively in natural language processing (NLP). They are adept at capturing long-term dependencies and intricate relationships within sequences, which is crucial for interpreting EHR data where past medical events can significantly influence future health outcomes. The transformer's self-attention mechanism allows it to weigh the importance of different parts of the sequence differently. This aspect is particularly beneficial in EHR data, where not all medical events have equal relevance to a patient’s current health status or future medical predictions. By focusing on more significant events in a patient's medical history, transformers can provide more accurate and personalized health predictions. Our project was mainly inspired by the following studies that used the Transformer model with electronic health record data:

\paragraph{\textbf{GCT model}}

Prior research in EHR data representation primarily utilized the Bag of Words (BOW) approach, treating each medical feature as an isolated entity. This methodology, however, led to significant information loss about the physician's decision-making process. The Graph Convolutional Transformer (GCT) model, presented by Choi et al.\cite{choi2020learning}, offers a robust solution to this issue by employing the Transformer architecture's self-attention mechanism. This mechanism effectively learns a hidden graphical structure, delineating the relationship between different EHR features, thereby overcoming the limitations of the BOW approach.

GCT represents EHR data as a two-dimensional matrix, where each cell indicates a connection between two features in the graph. This model's primary objective is to learn this hidden structure and utilize it for various predictive tasks, especially when explicit structural information is not available. The article highlighted that the learned hidden structure through GCT creates new embeddings, significantly enhancing the model's performance across various tasks. 

However, GCT has certain limitations. Its analysis is confined to single-time intervals—specifically, individual hospital visits—potentially overlooking the continuum of patient care. Furthermore, by dividing features into three categories—diagnosis, treatment, and lab tests—GCT's generality across diverse medical scenarios is somewhat constrained. Despite these drawbacks, GCT's approach in learning and utilizing hidden EHR structures demonstrates significant improvements in model performance for tasks like readmission and mortality prediction, marking a notable advancement in the application of Transformer models in the realm of EHR data.

\paragraph{\textbf{SARD model}}

Kodialam et al. \cite{kodialam2021deep} introduced a Transformer-based architecture known as SARD, innovatively combining embeddings for hospital visits, temporal embeddings, and a self-attention mechanism. This design deviates from the traditional positional embeddings, accommodating the non-uniform timing of hospital visits. In their study, they utilized a large dataset of administrative claims to predict end-of-life and surgeries in the next six months. We recently developed an extension of the SARD architecture, which we named STRAFE \cite{zisser}, with the goal of predicting time-to-event instead of fixed-time prediction and applied it to predict deterioration in chronic kidney disease.

One notable limitation of using claims data, and by extension the SARD model, is its exclusion of granular data from individual hospital visits, such as real-time monitoring signals (e.g., respiratory rate values). This gap highlights a potential area for model improvement in capturing finer details of patient care.

A unique aspect of the SARD model is its ability to discern connections between individual hospital visits, revealing a 'hidden structure' in the healthcare journey of a patient. However, it faces challenges in accurately representing scenarios with rapidly changing features, such as during an ICU stay. In such cases, the model may not fully capture the dynamic nature of a patient's condition, where variables like respiratory rate can fluctuate significantly over short periods.

\section{Framework Overview}

This RatchetEHR framework is primarily inspired by the SARD framework, which is used for claims data analysis. However, EHR data presents unique challenges compared to claims data, necessitating several key modifications in our approach.

First, unlike claims data which primarily consist of categorical features such as diagnosis and procedure codes, EHR data encompasses a wide array of numerical features. These include vital statistics like blood counts or respiratory rates, which are more akin to continuous signals than discrete categories.

Second, the temporal scope of the data differs markedly between these two domains. Claims data often span multiple visits over extended timeframes, offering a longitudinal view of a patient’s health history. In contrast, EHR data, particularly in the context of ICU stays, tends to be more focused, typically concentrating on a single hospital admission.

In light of these differences, we made several adjustments to the original SARD framework to better suit the specific requirements of ICU EHR data analysis. RatchetEHR introduces a transformer-based architecture that is specifically designed for the task of analyzing ICU EHR data. Central to our framework is the ability to process ICU stay information, represented as a 3-dimensional tensor, transformed into a contextualized format that can be used for downstream prediction, including BSI. 

\subsection{Data representation}
\label{data:representation}

In SARD, concept embedding plays a crucial role due to the categorical nature of the information related to hospital visits. Each piece of information is encoded as a 'word' and then transformed into a word embedding using techniques like Word2Vec. These word embeddings are aggregated to form a comprehensive visit embedding, summarizing the data from a single hospital visit. However, here we are modeling EHR data which predominantly consists of numerical information derived from charts and monitoring devices. This data can be best characterized as a series of signals, reflecting real-time physiological changes in patients. Directly applying concept embedding, as done in the SARD model, would lead to a substantial loss of critical information.

To effectively represent this dynamic and complex nature of EHR data, RatchetEHR adopts a distinct approach, inspired by the method outlined by Wang et al. \cite{wang2020trace}. Here, we represent the data in a structured form. We segment each patient’s ICU visit into discrete time intervals, termed as timeframes. Each timeframe encompasses the data relevant to its respective time interval. In constructing these time-frames, we categorize our data into two distinct types: numerical and categorical. Numerical data primarily includes readings from monitors and results from examinations – for example, respiratory rate and blood pressure. On the other hand, categorical data includes aspects such as diagnoses, which we represent using one-hot encoding vectors.

For each timeframe, we then use the numerical and categorical vectors to form a singular, comprehensive input embedding that represents that specific timeframe. This process is repeated for each interval, building a sequential representation of a patient's ICU stay. The result of this process is a two-dimensional matrix for each patient’s ICU visit, constructed by concatenating these time-frames in chronological order. This methodology allows RatchetEHR to maintain the integrity of both the numerical and categorical data, capturing the dynamic and complex nature of EHR data for each patient's ICU stay.

Formally, the framework splits each ICU stay into time frames (notated as TF) of $h$ hours. Each time frame $j$ of ICU stay $i$ can be viewed as the following vectors:

$$\vec{u_j^i} = ({u}_{j_1}^i, {u}_{j_2}^i, ..., {u}_{j_k}^i) \qquad
\vec{w_j^i} =  ({w}_{j_1}^i, {w}_{j_2}^i, ..., {w}_{j_m}^i)$$

Where ${v}_{j_k}^i$ is the median of all values of feature $k$ at time frame $j$ for ICU-stay $i$, and ${w}_{j_m}^i$ is an indicator for whether the code number $m$ occurred for ICU-stay $i$ at time frame $j$. Therefore, $({w}_{j_1}^i, {w}_{j_2}^i, ..., {w}_{j_m}^i)$ is a BOW (bag of words).

We will denote $l = k + m$ and use $||$ as the concatenation operation. 

Therefore,

$$\vec{v_j^i} =   \vec{u_j^i}\ ||\  \vec{w_j^i} \in \mathbb{R}^{l} $$

Each sample $i$ can be viewed as a 2-dimensional Tensor $\mathbf{W}^i \in \mathbb{R}^{p \times l}$, where $p$ is the number of time frames for the $i$ sample.
The rows of the tensor are the time frames of the ICU-stay $i$, which can be denoted more formally:
$$\mathbf{W}^i_j = \vec{v_j^i}$$

These samples are the input to the framework model.

\subsection{Model architecture}

Two primary distinctions set RatchetEHR apart from the SARD model. First, while SARD aggregates visit data into a singular sum of concept codes, obscuring potential hidden structures within a visit, RatchetEHR reincorporates the Graph Convolutional Transformer (GCT) component, as suggested by Choi et al \cite{choi2020learning}. This integration allows for a more nuanced understanding of the data.
 
\begin{figure}[htb] 
\includegraphics[scale = 0.5, trim={0 0.7cm 0 0},clip,width=6in]{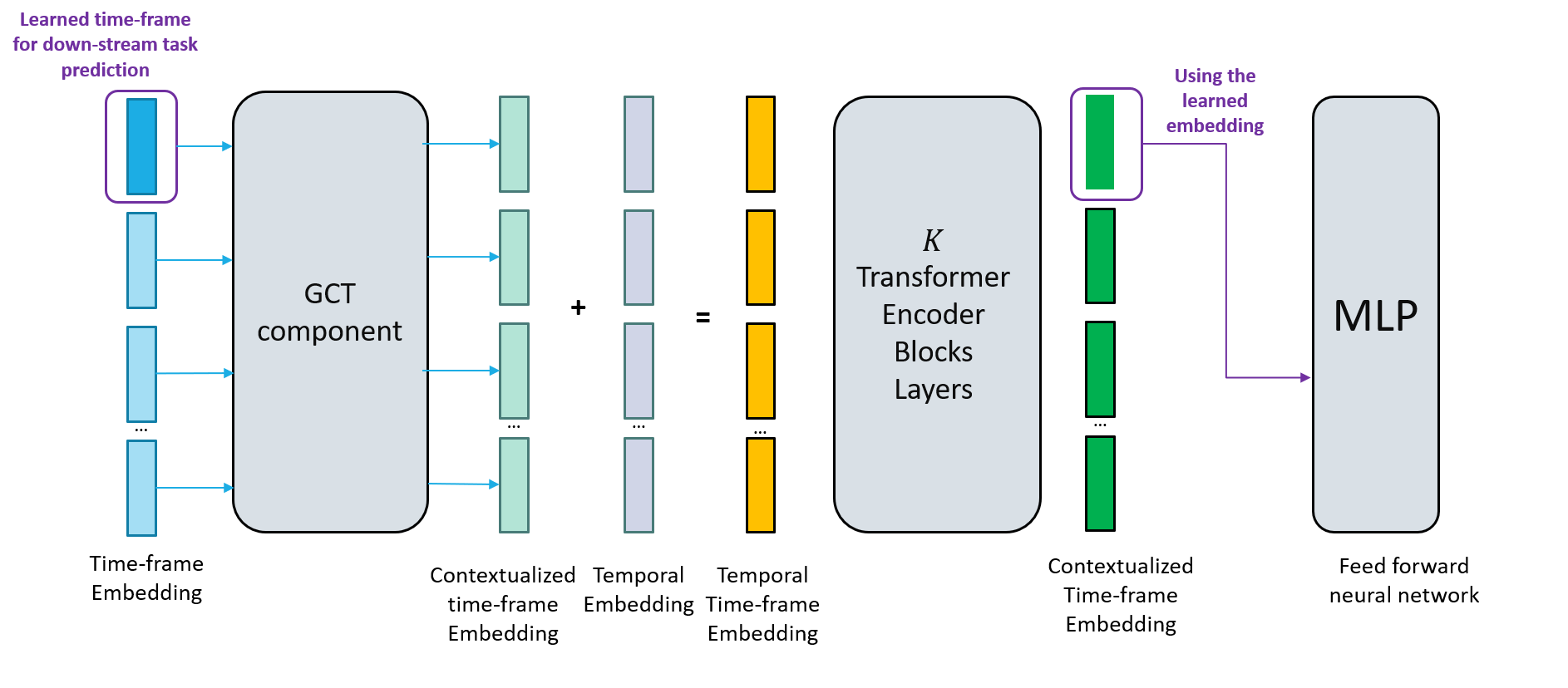}
\caption{\textbf{RatchetEHR architecture.} The architecture has three key components: Time Frame Embedding, Temporal Embedding, and Transformer Encoder. The integration of the Graph Convolutional Transformer (GCT) component is also depicted, highlighting its role in enhancing the ability of the model to identify hidden structural relationships within the data. Advanced methodologies such as Transfer Learning, Learned Time-frame Embedding, Focal Loss, and Child Tuning are incorporated to optimize the performance of the model, particularly in addressing challenges like limited sample sizes, class imbalance, and overfitting.}
\label{fig:modelarchitecturegct}
\end{figure}

The architecture of RatchetEHR, detailed in Fig. \ref{fig:modelarchitecturegct}, includes the following components:

\paragraph{\textbf{Time Frame Embedding}} 

As introduced in \ref{data:representation}, it is a 3-dimensional tensor input representation for the ICU data. It is composed of timeframes of $h$ hours of the EHR data, a signal-like format suitable for time-series analysis.

\paragraph{\textbf{Temporal Embedding}} 

Utilizing fixed positional encoding, maintains the chronological order of the timeframes within the ICU stay, crucial for preserving the sequential nature of the data. Contrary to Kodialam et al. \cite{kodialam2021deep}, where the visits do not occur regularly, the time frames are regular. Therefore, we used the fixed positional encoding introduced in the original Transformer article \cite{vaswani2017attention}:
$$PE_{(pos, 2i)} = sin(\dfrac{pos}{10000 ^ {\frac{2i}{l}}}) \qquad
PE_{(pos, 2i+1)} = cos(\dfrac{pos}{10000 ^ {\frac{2i}{l}}})$$

Where $pos$ is the position and $i$ is the dimension.

\paragraph{\textbf{Transformer Encoder}} To effectively handle the sequential and time-variant nature of EHR data, we utilize the Transformer model. It is adept at creating contextualized embeddings that correlate with other time-frames. This is critical as most features at a certain time are dependent on their previous values. We include $K$ transformer encoder layers, as delineated in Vaswani et al. \cite{vaswani2017attention}, to output contextualized time frame embeddings.

\paragraph{\textbf{Learned Time-frame Embedding}} To mitigate the risk of overfitting in tasks with limited EHR data, we incorporated a learned time-frame embedding for each input. This concept, inspired by the BERT [CLS] token \cite{BERT}, aids in reducing the parameter count for downstream task predictions by encapsulating essential information within a trainable parameter.

\paragraph{\textbf{MLP}} A feed-forward neural network that transforms the learned timeframe embedding into a probability that indicates the class of the sample.

\subsection{Model Refinements}

To refine the RatchetEHR architecture, we integrated several advanced methods, addressing key challenges encountered during the experimentation with our SARD-inspired base model. A primary concern was the tendency of Transformer models, due to their large parameter count, to overfit, especially when dealing with limited sample sizes typical of EHR prediction tasks. Despite this, their strength in processing the sequential nature of EHR data is undeniable.

\paragraph{\textbf{Transfer Learning}}
We adopted a dual-stage approach, akin to the BERT model's methodology \cite{BERT}, consisting of pre-training and fine-tuning stages. The pre-training phase employs a masking task on the extensive EHR data samples to initialize the model's weights, which is crucial for enhancing performance in downstream tasks. The architecture is augmented with a linear layer, projecting the contextualized time frame embeddings back into the input initial representation.

\paragraph{\textbf{GCT Component}}
We integrated a Graph Convolutional Transformer (GCT) component into RatchetEHR to address another baseline model limitation: the lack of consideration for inter-feature relationships. By modeling these relationships as a hidden graphical structure, the GCT component enhances the robustness of time-frame embeddings in \cite{choi2020learning}. Trained across numerous time-frame embeddings, this method not only boosts the model's ability to discern hidden structures but also helps in circumventing overfitting.

\paragraph{\textbf{Sampler and Focal Loss}}
To address class imbalance and its resultant prediction bias and overfitting, we implemented a weighted sampler for oversampling, creating more balanced mini-batches for training. The class weight, $w_i$, is inversely proportional to the class frequency, $n_i$. Additionally, focal loss is employed, focusing the model's learning on more challenging examples, a technique widely used in object recognition tasks with high class imbalance.

\paragraph{\textbf{Child Tuning}}
To further enhance the model's efficiency and reduce overfitting, Child Tuning, as described in \cite{ChildTuning}, is employed. This method limits the training to only the most relevant parameters identified through Fisher information, which assesses the sensitivity of the model to changes in each parameter. This selective training approach streamlines the model, focusing on parameters most critical to the task at hand.

\vspace{-0.5cm} 

\vspace{-1cm} 

\section{Experiments}

\subsection{Datasets}

We utilized the publicly available dataset MIMIC-IV which contains deidentified EHR data from 73,181 patients and 299,712 ICU stays at Beth Israel Deaconess Medical Center between 2008-2019. MIMIC-IV provides a rich source of EHR information, including vital signs, medication records, laboratory results, and patient demographics, acquired by routine clinical care, monitors and more. We utilized the ICU module of the MIMIC-IV dataset. This module provides detailed information about individual patient visits to the ICU, including subject ID, start and end times of the ICU stay, and various medical measurements and events recorded during the stay. The data was stored in a PostgreSQL database.

To facilitate data extraction and manipulation, RatchetEHR employs SQLAlchemy, a Python-based SQL toolkit, to generate patient cohorts and retrieve relevant feature information from the PostgreSQL database. This extraction process focuses on critical ICU metrics such as vital signs, medication records, laboratory results, and patient demographics, ensuring a comprehensive dataset for analysis.

Data preparation involved cleaning processes to handle missing values and outliers in the dataset. We employed linear interpolation provided by pandas framework to address gaps in the data and utilized established medical thresholds to identify and rectify out-of-range values \cite{mimic-extract}.

To reduce the space of ICD-10 codes we utilized a mapping of ICD-10 codes to diseases as provided by \cite{kuan}, which helped streamline the dataset, making it more manageable and conducive for our analysis.

\subsection{Prediction Task}
\label{prediction_task}

Bloodstream Infection (BSI) is a critical condition that significantly impacts ICU patients, resulting in prolonged hospital stays, life-threatening complications, and notably high morbidity and mortality rates exceeding 30\%, \cite{10.1001/archinte.1995.00430060050006}. The standard diagnostic procedure for BSI involves a blood culture test, which typically requires one to two days to yield results. This delay is critical, considering the rapid progression of BSI and its severe consequences. Prompt detection and immediate antibiotic treatment are crucial for reducing the associated high morbidity and mortality rates, yet early-stage detection remains a challenge for physicians.

Our objective was to train RatchetEHR and other machine-learning models to predict BSI in patients who underwent a blood culture test and remained in the hospital for a minimum of two days post-test. This approach aligns with several existing studies aimed at forecasting BSI risk. Our architecture was specifically tested for its effectiveness in predicting BSI.

For cohort building, we replicated the method used by Roimi et al. \cite{roimi}, adhering to the guidelines set by the Center for Disease Control and Prevention (CDC)/National Health Safety Network (NHSN). We identified patients with BSI by detecting common commensal bacteria related to BSI, as listed in the NHSN organism tab. Patients showing growth of these bacteria in blood culture tests were labeled as positive for BSI. The detection was based on blood collection entries in the measurement table where the measurement attribute was blood culture.

\begin{figure}[htb]
\centering
\includegraphics[scale = 0.5]{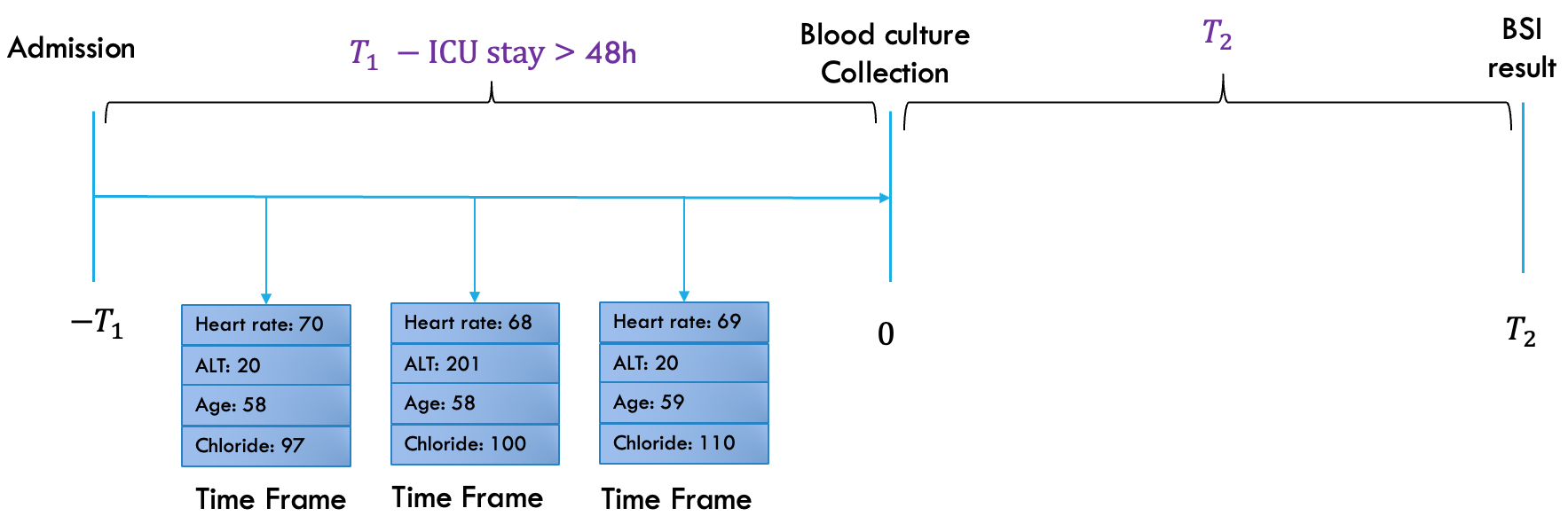}
\caption{\textbf{Prediction task.} The study design for each patient in the hospital. 0 is the index date which is the time of the blood culture collection. $T_1$ is the number of hours of the admission to the hospital before the blood culture test. During this interval, we collected the features. In the interval [0, $T_2$], we refrained from collecting data to prevent data leakage, as this is the period between the test and the results (the interval time is higher than 24 hours).
}
\label{fig:study-design}
\end{figure}

Our selection criteria for ICU stays focused on cases where blood collection occurred more than 48 hours after admission. This criterion aimed to exclude patients who were admitted to the ICU primarily for surgical reasons, as they are generally not at risk for BSI. The study design, illustrated in Fig. \ref{fig:study-design}, presents the timeline for each patient's hospital stay relative to the blood culture collection.

\vspace{-0.5cm} 

The inclusion criteria included undergoing blood culture test, duration of at least 48 hours at the ICU before blood culture test, and admission to the MICU (Medical ICU), SICU (Surgical ICU) and TSICU (Trauma ICU). In total, our cohort comprised 12,139 ICU stays, of which 538 were identified with BSI, representing a prevalence of 4.4\%, highlighting the highly imbalanced nature of the dataset. It is important to note that BSI is sometimes treated without conclusive laboratory results and may not always be consistently coded in EHR data. This could potentially lead to an under-prediction of true BSI cases, which is a major limitation of using EHR data for this type of analysis.

Each sample consisted of at most 30 timeframes, with each timeframe encompassing four hours of data. This setup effectively captures a comprehensive timeline of five days leading up to each blood culture test. To ensure robustness in our model evaluation and to prevent data leakage associated with the year of ICU stay, we strategically split the dataset into training-validation and test sets based on the year of the ICU admission. Specifically, ICU stays from the years 2008 to 2017 were allocated to the training-validation set, while those from 2017 to 2019 were designated for the testing set. This split was chosen to ensure a sufficient sample size for training while maintaining a temporal separation between the training and testing data, which can help assess the model's performance on future, unseen data.

Given the highly imbalanced nature of the task, with a significantly lower prevalence of BSI cases, we opted to evaluate our model using the AUC-ROC score, a metric less sensitive to class imbalance. Additionally, considering the relatively small size of the sample dataset, we anticipated a high variance in performance metrics due to the initial split of the data into training, validation, and test sets. To account for this variability and to ensure a comprehensive evaluation, we conducted 10 different experimental runs for each model. In each iteration, the training, validation, and test sets were randomly selected, providing a thorough and varied assessment of the model's performance across multiple splits of the dataset.

\subsection{Benchmark Models}

\paragraph{\textbf{RNN}}
We employed an RNN model inspired by the architecture and hyperparameters detailed in \cite{zachery}. The model uses Tanh activation and consists of two RNN layers with batch normalization. The final hidden state of the RNN feeds into a linear layer, projecting it to the probability of BSI occurrence in patients.

\paragraph{\textbf{LSTM}}
Adapting the RNN model architecture, the LSTM variant replaces RNN layers with LSTM layers while maintaining the same hyperparameters. LSTM models are particularly adept at processing longer sequences, offering an improvement over traditional RNNs. However, it is important to note that LSTM models process inputs sequentially.

\paragraph{\textbf{LSTM / RNN + Focal Loss}}
We also evaluated LSTM and RNN models trained with focal loss. Focal loss is instrumental in directing the model's attention towards more challenging, incorrectly classified examples, thereby enhancing model performance on complex cases.

\paragraph{\textbf{XGBoost}}
For the XGBoost classifier, as provided by the dmlc XGBoost package, we adapted our input data to fit the classifier’s requirements. Given that XGBoost processes two-dimensional data, and our input is a three-dimensional matrix, we reduced the dimensionality by computing the median of the last eight timeframe embeddings (covering four hours) before the blood culture collection. Hyperparameter tuning was conducted using a randomized cross-validation search over 10,000 iterations, utilizing the scikit-learn framework.

\paragraph{\textbf{Random Forest}}
In the case of the RandomForest classifier, hyperparameter optimization was achieved through a randomized cross-validation search with 10,000 iterations. We employed balanced class weights to address the class imbalance inherent in BSI prediction data.

\paragraph{\textbf{RatchetEHR variations}}
Several variations of the RatchetEHR model were tested by modifying or removing specific components. For the transfer learning (TL) approach in the pretraining stage, the AdamW optimization algorithm was utilized, along with parameters such as a batch size of 32, dropout rate of 0.1, learning rate of $10^{-4}$, and weight decay of 0.2. The fine-tuning stage hyperparameters were based on the BERT article \cite{BERT}, including the warmup method for learning rate adjustment. The learning rate gradually increases during the initial $m$ steps of the optimization algorithm and then decreases linearly. A smaller batch size of 17 was used, along with a dropout of 0.5, an initial learning rate of $\alpha\cdot10^{-3}$, and a weight decay of $\beta\cdot0.3$.

\subsection {Prediction Performance}

\begin{figure}[h] 
\subfloat[]{\includegraphics[width = 3in]{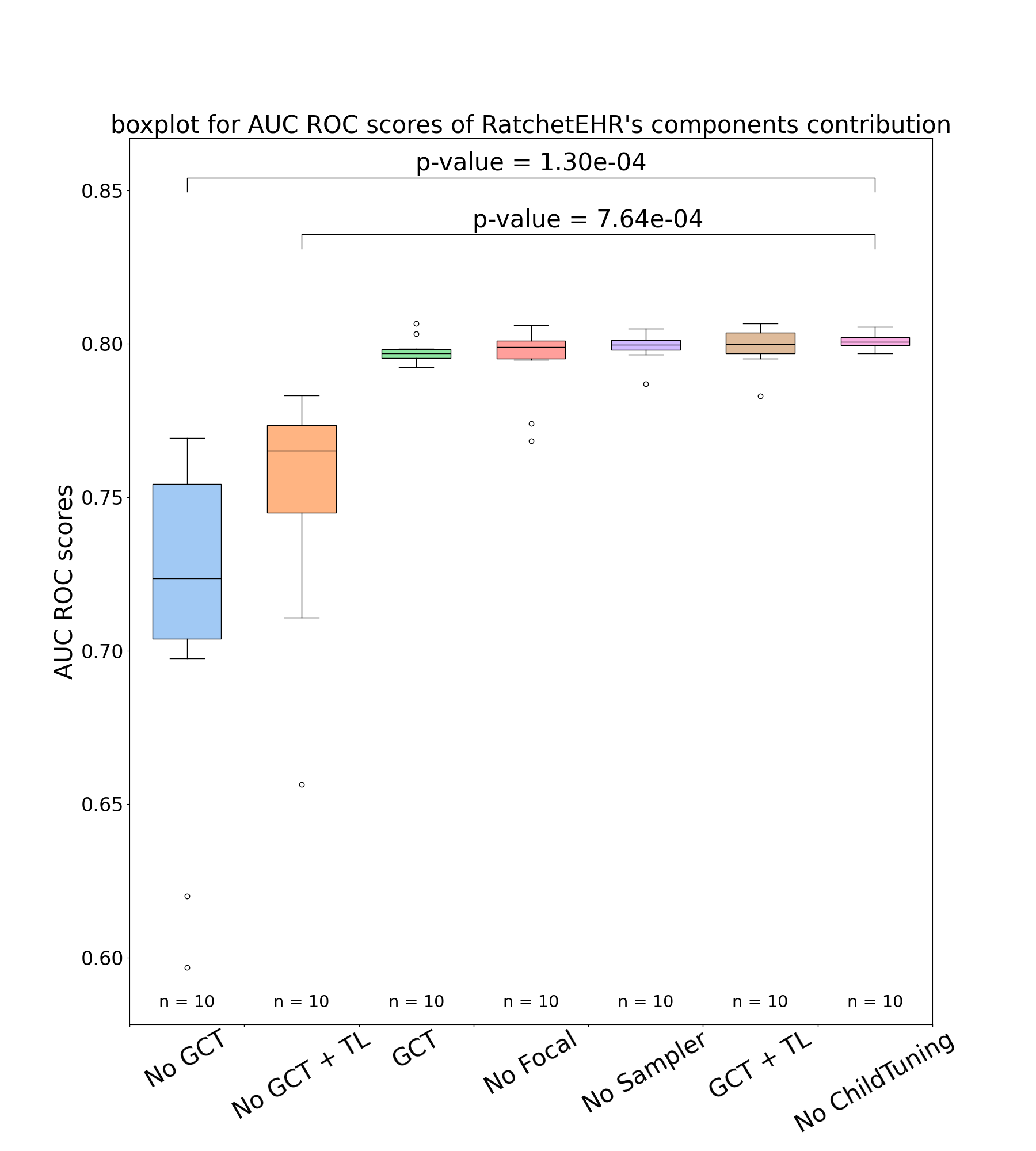}} 
\subfloat[]{\includegraphics[width = 3in]{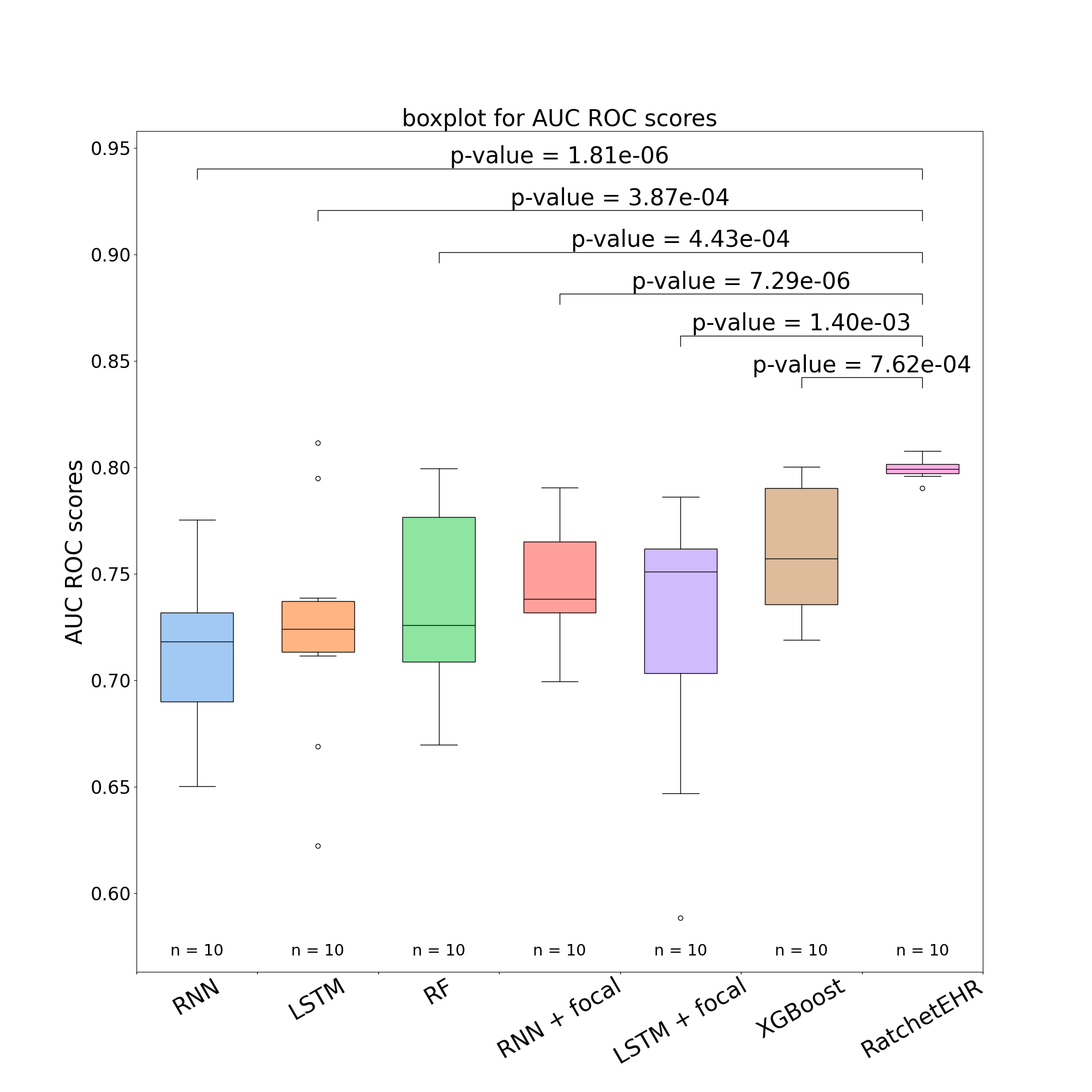}}
\caption{\textbf{Evaluation of RatchetEHR performance. A.} Boxplots show AUC-ROC on the test sets in the 10 iterations. We compared different variations of our architecture to showcase the relative contribution to performance of each component. Values were compared using t-Test. The GCT component provided a significant boost to the performance. TL: transfer learning approach. \textbf{B.} Boxplots show AUC-ROC on the test sets in the 10 iterations. We compared different algorithms to the full version of RatchetEHR. Values were compared using t-Test. RF: Random Forest.
}
\label{fig:Benchmark}
\end{figure}

We first assessed the contribution of the different components used in the RatchetEHR architecture. The analysis revealed that the incorporation of the GCT component substantially enhanced the performance of the prediction (Fig. \ref{fig:Benchmark}. [A]). This improvement can be attributed to the ability the GCT component to train on a vast number of timeframe embeddings, each representing a single timeframe. This extensive training leads to more robust and contextually enriched input embeddings that effectively capture the hidden graphical structure of EHR data features. Consequently, this contributes to mitigating the challenges posed by the class imbalance and the limited size of the dataset.

In the model variations lacking the GCT component, the transfer learning approach boosts performance and reduces the variance in the test set. This suggests that transfer learning is particularly effective in refining the model's accuracy in the absence of the GCT component. Though small differences, the top-performing architecture was the one that included all components, including GCT, focal loss, sampler, transfer learning, but without child tuning (average AUC-ROC = $0.8 \pm 0.002$). We suggest that this might be due to the substantial drop in the number of parameters that the child tuning masked and therefore were not updated.

Comparative analysis with other models highlights that while traditional models like RNN and LSTM show varying degrees of effectiveness, they are somewhat limited, especially in small datasets (Fig. \ref{fig:Benchmark}. [B]). The application of focal loss function, commonly used in object detection tasks, shows improvements in these models, but not drastically. XGBoost, known for its proficiency with tabular data, shows higher performance, yet it is outperformed by RatchetEHR. This superior performance of RatchetEHR is largely due to its ability to process raw data through the transformer model, which efficiently discovers hidden structures and relationships within the time-frame embedding features and among the time-frames themselves.

\vspace{-0.8cm} 

\vspace{-0.8cm} 

\section{Explainability}

Machine learning models are prone to bias, confounders and other issues. This leads to mistrust among the users, especially in the healthcare domain, where the models output affect the patients' quality of life. Many studies were conducted to provide insight to the model's decision making, to enhance the users' assurance on it. To provide a deeper understanding of how our RatchetEHR model arrives at its predictions, we utilized SHAP values. SHAP values offer an insightful way to interpret complex machine learning models, as explained in \cite{SHAP}. Here, we generated SHAP summary plots for 100 randomly selected ICU stays. Given the extensive number of features involved in each ICU stay, we focused on optimizing performance without compromising the depth of our analysis. We utilized the GradientExplainer, a component of the SHAP framework. This tool efficiently approximates SHAP values using expected gradients, as detailed in \cite{sundararajan2017axiomatic}. This approach enabled us to maintain computational efficiency while still providing rich, interpretable insights into the features driving the model’s predictions.

\begin{figure}[h]
\subfloat[]{\includegraphics[width = 3in]{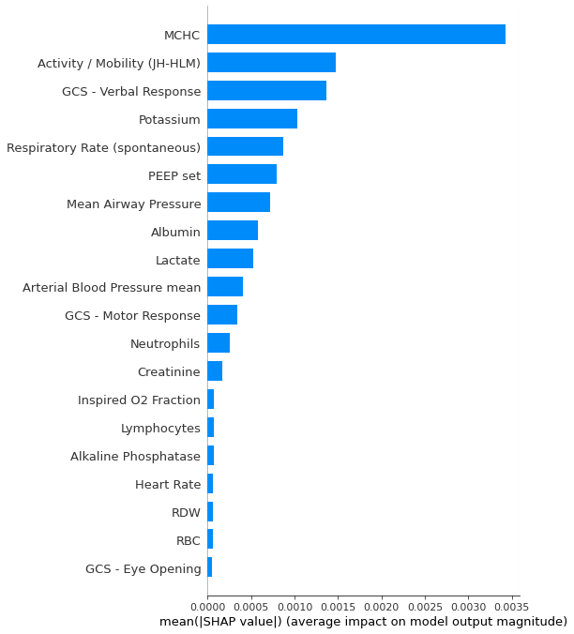}} 
\subfloat[]{\includegraphics[width = 3in]{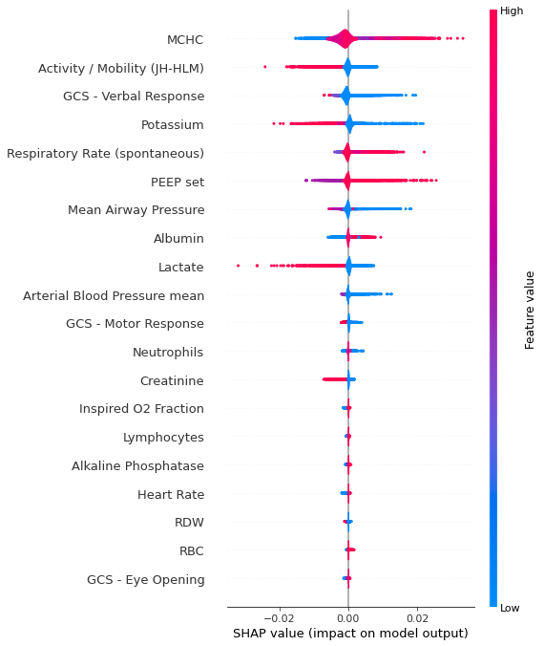}}
\caption{\textbf{Explainability of the model. A.} SHAP summary bar plot, displaying the importance of each feature. \textbf{B.} SHAP summary violin plot, illustrating how feature values affect the model's predictions.}
\label{some example}
\end{figure}

This analysis revealed that the most important feature in our prediction of BSI was Mean Corpuscular Hemoglobin Concentration (MCHC) (Fig. \ref{fig:Benchmark}). Typically, MCHC is a measure of the average concentration of hemoglobin in a person's red blood cells, used primarily to diagnose and monitor conditions related to red blood cell health, such as anemia. This finding aligns with previous research where MCHC was identified as a relevant factor in BSI, as noted in studies by Roimi et al. \cite{roimi} and Zoabi et al. \cite{zoabi2021}, suggesting a potential, yet not fully explored, link between MCHC levels and BSI. Other features that our analysis underscored include the Glasgow Coma Scale (GCS) scores for verbal and motor responses, which echo the findings in Roimi et al. \cite{roimi} and Mahmoud et al. \cite{mahmoud}. Additionally, variables such as Albumin levels, respiratory rates, creatinine, and heart rate were also identified as significant, consistent with observations in the studies by Mahmoud et al. \cite{mahmoud} and Zoabi et al. \cite{zoabi2021}. It is important to note that these correlations, while statistically significant in the context of the model, may not directly imply a causal relationship. Rather, it could reflect complex interplays in the patient's health status, where alterations in blood measurement levels coincide with factors that contribute to the susceptibility or onset of BSI.

\section{Discussion and Conclusions}

We presented here a complete framework for modeling EHR data of hospitalizations using a transformer-based architecture. We show that this framework provides superior performance over other state-of-the-art machine-learning approaches. This architecture is adept at effectively processing sequential EHR data, a crucial aspect given the temporal nature of medical records and their importance in clinical decision-making. This capability is particularly vital in predicting conditions like BSI, where the timing and evolution of patient data points are key indicators of the patient's health trajectory.

A pivotal aspect of RatchetEHR is the integration of the GCT component. By leveraging this component, RatchetEHR can uncover the hidden structural relationships within the data, crucial for understanding complex clinical scenarios. The GCT component notably enhances the ability to process and interpret each timeframe of EHR data. This capability is instrumental in the superior performance of our framework. Other components added to the architecture, including the focal loss, Sampler, and ChildTuning did not result in significant performance improvements, however, they were incorporated to address specific challenges such as class imbalance and overfitting. Future work could explore simplifying the model architecture to strike a balance between complexity and performance.

While complex, we show here that is it possible to extract feature importance from the model and provide the much-needed explainability. This aspect is highly valuable in clinical settings, where understanding the 'why' behind a model's prediction is as crucial as the prediction itself. This transparency allows clinicians to trust and effectively utilize AI-driven insights in their decision-making process.

It is important to consider potential limitations and biases inherent in EHR data. Inconsistencies in data collection, documentation, and coding practices across different healthcare systems may impact the model generalizability. For example, we are probably under-predicting cases of BSI due to inconsistent administrative coding and treatment without conclusive laboratory results. Future work could involve validating the model's performance on independent EHR datasets to assess its robustness and transferability, in addition to a prospective study in real-world scenarios. 

In conclusion, this study contributes to the field of medical informatics by introducing an innovative approach to EHR data analysis and opens up new possibilities for future research to further enhance AI capabilities in healthcare.

\paragraph{\textbf{Data and code availability}}\label{dataavailability}

MIMIC-IV was downloaded from the PhysioNet project: (\url{https://physionet.org/content/mimiciv/2.2/}. The code developed in this study is available at \url{https://github.com/OrtalHirszowicz/RatchetEHR}.

\paragraph{\textbf{Acknowledgments}}

DA is supported by the Azrieli Faculty Fellowship. We thank Joseph Bingham for constructive review of the manuscript.

\paragraph{\textbf{Funding Statement}}

This research received no specific grant from any funding agency in the public, commercial or not-for-profit sectors

\paragraph{\textbf{Competing Interests Statement}}

DA reports consulting fees from Carelon Digital Platforms. OH has no competing interest.

\bibliographystyle{unsrt}
\bibliography{sample}

\end{document}